\documentclass[conference]{IEEEtran}
\IEEEoverridecommandlockouts
\usepackage{cite}
\usepackage{amsmath,amssymb,amsfonts}
\usepackage{algorithmic}
\usepackage{graphicx}
\usepackage{textcomp}
\usepackage{xcolor}
\usepackage{breqn}
\usepackage{booktabs} 
\usepackage{multirow}
\usepackage{array}
\usepackage{threeparttable}
\usepackage{hyperref}
\usepackage[linesnumbered,ruled,lined]{algorithm2e}
\usepackage{subfig}
\newcolumntype{P}[1]{>{\centering\arraybackslash}p{#1}}

\def\BibTeX{{\rm B\kern-.05em{\sc i\kern-.025em b}\kern-.08em
    T\kern-.1667em\lower.7ex\hbox{E}\kern-.125emX}}

\begin{document}

\title{Autonomous Cross Domain Adaptation under Extreme Label Scarcity}

\author{Weiwei Weng, Mahardhika~Pratama,~\IEEEmembership{Senior Member,~IEEE}, Choiru Za'in, Marcus de Carvalho, \\Rakaraddi Appan, Andri Ashfahani, Edward Yapp Kien Yee
\thanks{W. Weng and M. Pratama share equal contributions. W. Weng, A. Ashfahani, M. de Carvalho, R. Appan are with the School of Computer Science and Engineering, Nanyang Technological University, Singapore. M. Pratama is with the academic unit of STEM, University of South Australia, Adelaide, Australia. C. Za'in is with the school of IT, Monash University. E. Y. Kien Yee is with Singapore Institute of Manufacturing Technology, A*Star, Singapore. Majority of the work was done when M. Pratama was with SCSE, NTU, Singapore. E-mail: weiwei.weng@ntu.edu.sg; dhika.pratama@unisa.edu.au;choiru.zain@monash.edu;marcus.decarvalho@
ntu.edu.sg;appan001@e.ntu.edu.sg;andriash001@e.ntu.edu.sg;edward\_yapp@
simtech.a-star.edu.sg.}}

\maketitle

\begin{abstract}
 A cross domain multistream classification is a challenging problem calling for fast domain adaptations to handle different but related streams in never-ending and rapidly changing environments. Notwithstanding that existing multistream classifiers assume no labelled samples in the target stream, they still incur expensive labelling cost since they require fully labelled samples of the source stream. This paper aims to attack the problem of extreme label shortage in the cross domain multistream classification problems where only very few labelled samples of the source stream are provided before process runs. Our solution, namely Learning Streaming Process from Partial Ground Truth (LEOPARD), is built upon a flexible deep clustering network where its hidden nodes, layers and clusters are added and removed dynamically in respect to varying data distributions. A deep clustering strategy is underpinned by a simultaneous feature learning and clustering technique leading to clustering-friendly latent spaces. A domain adaptation strategy relies on the adversarial domain adaptation technique where a feature extractor is trained to fool a domain classifier classifying source and target streams. Our numerical study demonstrates the efficacy of LEOPARD where it delivers improved performances compared to prominent algorithms in 15 of 24 cases. Source codes of LEOPARD are shared in \url{https://github.com/wengweng001/LEOPARD.git} to enable further study. 
\end{abstract}

\begin{IEEEkeywords}
Multistream Classification, Transfer Learning, Data Streams, Incremental Learning, Concept Drifts
\end{IEEEkeywords}

\section{Introduction}
	\noindent\textit{Background}: Multistream classification problems \cite{MSC} portray a classification problem across many streaming processes running simultaneously but independently. Each streaming process features different but related characteristics to be handled by a single model having a stream-invariant trait. That is, each stream suffers from \textbf{the domain shift problem} in which they follow different distributions. Multistream classification problem also considers \textbf{the issue of labelling cost} where the ground truth access is only provided in the source stream while leaving the target stream with the absence of any labelled samples. Unlike the traditional domain adaptation problems, the multistream problem deals with continuous information flows which must be handled in the fast and sample-wise fashion. Another typical problem is \textbf{the asynchronous drift problem} which distinguishes itself from the conventional single stream problem. The asynchronous drift problem refers to independent drifts between source and target streams taking place at different time points. The multistream classification problem distinguishes itself from the online transfer learning problem \cite{OTL} in which both source and target domains are streaming in nature whereas the online transfer learning problem assumes a static source domain although it considers a streaming problem of the target domain. The underlying goal of the multistream classification problem is to build a predictive model $f(.)$ which simultaneously performs the unsupervised domain adaptation as well as addresses the issue of data streams. Notwithstanding the recent progress of the multistream classification area, most works are designed from a single domain perspective in which both source and target streams are drawn from the same feature space. In addition, existing solutions incur expensive labelling cost because they require a full supervision of the source stream. 
	
	\noindent\textit{Practical Scenario}: This paper puts into perspective a cross-domain multistream classification problem under \textbf{extreme label scarcity} where, unlike conventional multistream classification problems, the source stream and the target stream are generated from different feature spaces but share the same target attributes. The extreme label scarcity issue presents in the fact that no label is provided for the target stream while only few labelled samples are made available in the source stream during the warm-up phase. That is, an operator is only capable of labelling \textbf{few prerecorded samples of the source stream} while leaving the rest of data samples of the source steam unlabelled. The practical scenario of this problem is seen in the condition monitoring problem involving different machines. Instead of building a machine-specific model for monitoring purposes, a single machine-invariant model is constructed thereby saving significant developmental costs because data collection, annotation and preprocessing do not have to be repeated for each machine. Nevertheless, this task is challenging because data samples captured by sensors are streaming in nature. Different machines are installed with different sensors or of different types thereby producing different feature spaces while having different sampling rates leading to different batch sizes. Process's deviations due to tool wear or any other external influencing factors occur independently to each machine at different time points leading to drifting data distributions in each machine with different rates, magnitudes, types. The issue of labelling cost occurs because visual inspections leading to interruption of machine operations are necessitated to annotate data samples. It hinders the labelling process during the process runs. The labelling process is possible to be done only for prerecorded samples to avoid frequent stoppages of machine operations. 
	
\begin{figure}[!t]
	    \centerline{\includegraphics[scale=0.4]{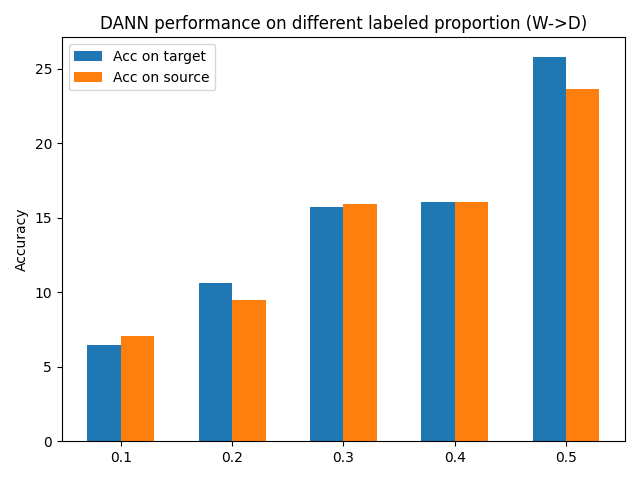}}
	    \caption{DANN performance on Office31 (D$\rightarrow$W) different label proportions of source streams leaving the target stream unlabelled.}
	    \label{fig:dann label}
	\end{figure}

	We visualize the significance of label's scarcity in the context of domain adaptation in Fig. \ref{fig:dann label} where DANN \cite{DANN} is evaluated under different label proportions of source streams. Our numerical results are produced in the office31 problem (Webcam $\rightarrow$ DSLR) using five label ratios: $10\%,20\%,30\%,40\%,50\%$. It is observed that DANN's performances are significantly compromised with reductions of label proportions, less than $10\%$ accuracy on source and target streams under $5,10\%$ label proportions. That is, its accuracy on source and target streams consistently slips. 
	
	\noindent\textit{Our Contribution}: Learning Streaming Process from Partial Ground Truth (LEOPARD) approach is proposed in this paper and resolves the cross-domain multistream classification problems under extreme label scarcity. LEOPARD is developed under the framework of a flexible deep clustering network where it features an elastic and progressive network structure to handle changing data distributions. That is, hidden nodes, hidden layers and hidden clusters are self-evolved in respect to the asynchronous drift problem in both source and target streams. The learning process of LEOPARD aims to achieve two objectives under shared network parameters: clustering-friendly latent space and cross domain alignment in which it minimizes three loss functions. The reconstruction loss functions as the nonlinear dimension reducer where it projects input samples into a low dimension and establishes a common latent space between the source stream and the target stream. It is achieved by the stacked autoencoder (SAE) performing nonlinear mapping. The second component is the clustering loss creating a clustering-friendly latent space preventing the trivial solution. The cross domain adaptation loss is meant to induce the domain alignment and utilizes the adversarial domain adaptation approach \cite{DANN}. This strategy relies on a domain classifier to classify the origin of data samples and a feature extractor. The feature extractor and the domain classifier compete to each other thus resulting in domain-invariant representations. LEOPARD does not call for any labelled samples for its updates and few prerecorded labelled samples of the source stream are only used to establish the class-to-cluster relationship. 
	
	This paper presents four major contributions: 1) it proposes a new problem, namely cross domain multistream classification problems under extreme label scarcity; 2) an algorithm, namely LEOPARD, is developed to address the issue of label's scarcity in the cross-domain multistream classification problem; 3) a joint optimization problem is formulated to attain the clustering-friendly latent space as well as the domain alignment such that the target stream can be predicted accurately with very few labels of the source stream and no labels of the target stream; 4) the source code of LEOPARD along with all datasets are made public in \url{https://github.com/wengweng001/LEOPARD.git} to enable further study. Our numerical study has substantiated the efficacy of LEOPARD in handling the issue of extreme label scarcity in the cross domain multistream classification problems. It delivers highly competitive performances compared to prominent algorithms.

\section{Related Works}
	\noindent\textit{Multistream Classification}: The area of multistream classification has attracted growing research interests as observed by the number of works published in the literature. A pioneering work is proposed in \cite{MSC} using the kernel mean matching (KMM) method as a domain adaptation technique combined with a drift detection method to detect the concept drift in each domain. Considering high computational complexity and memory demand of \cite{MSC}, FUSION is proposed in \cite{FUSION} where the KLIEP method is implemented for domain adaptation while a density ratio method is designed for detecting the asynchronous drifts. MSCRDR is put forward in \cite{MSCRDR} and uses the Pearson divergence method for domain adaptation. Recently, a deep learning algorithm, namely ATL, is proposed to solve the multi-stream classification problem using the encoder and decoder structure under shared parameters coupled with the KL divergence method for domain adaptation \cite{ATL}. ATL characterizes an inherent drift handling aptitude with a self-evolving network structure. MELANIE is proposed in \cite{MELANIE} to handle the multi-source multistream classification problem. This work is extended in \cite{MARLINE}. Another solution of multi-source multistream classification is offered in \cite{AOMSDA} where the CMD-based regularization is integrated. The problem of multisource unsupervised domain adaptation under both homogeneous and heterogeneous settings are discussed in \cite{Liu2021MultisourceHU}. The area of multi-stream classifications deserves an in depth study due to at least two reasons: 1) these approaches are designed for a single domain problem where both source and target streams share the same feature space (domain). To the best of our knowledge, there exists only one work in the literature handling the cross-domain multistream classification problem \cite{COMC} using the empirical maximum mean discrepancy for domain adaptation. However, this approach is based on a non-deep learning solution relying on a simple linear projection for feature transformation, prone to trivial solutions; 2) Although these approaches rely on the unsupervised domain adaptation approaches where no label is offered for the target stream, full annotations are required for the source stream. On the other side, the multistream classification problem distinguishes itself from the online transfer learning problem \cite{OTL} assuming a fixed and static source domain. Hence, the problem of asynchronous drift problem is absent in the online transfer learning problem. 
	
	\noindent\textit{Semi-Supervised Transfer Learning}: The issue of labelling cost has attracted research interest in the transfer learning community. In \cite{CLARINET}, the notion of complementary labels incurring less expensive labelling cost than true class label is implemented. Dual deep neural networks is designed in which one focuses on complementary labels while another handles the domain adaptation. \cite{HEUDA} concerns on reductions of the labelling cost in the heterogeneous domain adaptation problem usually calling for some labelled samples of the target domain. Another effort is devoted to reduce the labelling cost in \cite{UOSDA} where it concerns on an open set domain adaptation where the target domain contains unknown classes. The use of noisy labels for unsupervised domain adaptation has been investigated in \cite{Liu2019ButterflyRO}. Our work differs from these works in two aspects: 1) LEOPARD handles the situation of cross-domain multistream classification under extreme label scarcity. That is, labelled samples are only revealed for the source stream during the warm-up period while no labelled samples for both streams are given for model updates during the process runs; 2) The learning approach is designed for the stream learning scenario. 
    
\section{Problem Formulation}
	Suppose that $D_S,D_T$ stand for the source and target domains respectively. The goal of domain adaptation is to solve a classification problem of the target domain $D_T$ without any labels by transferring knowledge base from the source domain $D_S$ where there exist some labelled samples. Referring to \cite{ben2010theory}, the generalization error of the target domain is upper bounded by how good a model $f(.)$ learns the source domain and the discrepancies between two domains:
	\begin{dmath}\label{bound}
	    \epsilon_T(f)\leq\underbrace{\epsilon_S(f)}_{1^{st}}+\underbrace{d_1(D_S,D_T)}_{2^{nd}}\\+\underbrace{\min{(E_{D_S}(f_S,f_T),E_{D_T}(f_S,f_T))}}_{3^{rd}}
	\end{dmath}
	 where the first term is the source error, the second term is divergence between the two domains and the last term is the difference in labelling function between the two domains which should be small \cite{ben2010theory}. Direct minimization of the divergence is a challenging task due to a lack of correspondences between data samples of the two domains. For streaming data, it leaves a major challenge because the divergence measure works with a finite number of samples. 
	 
	 A cross-domain multistream classification problem under extreme label scarcity is defined as a classification problem of two independent streaming data $B_1^S,B_2^S,...,B_{K_S}^S$ and $B_1^T,B_2^T,...,B_{K_T}^T$ termed as a source stream and a target stream respectively where $K_S,K_T$ are respectively the number of source stream and target stream unknown in practise. $B_{k_s}^S,B_{k_t}^T$ are drawn from the source domain $D_S$ and the target domain $D_T$ respectively. \textbf{Extreme label scarcity} is perceived in the limited access of ground truth where only prerecorded samples of the source stream $B_0^S=\{x_i^S,y_i^S\}_{i=1}^{N_m}$ are labelled while no label is provided during the process runs $B_{k_s}^S=\{x_{i}^S\}_{i=1}^{N_{S}}$. $N_m, N_S$ denote the number of prerecorded data samples of the source stream and the size of the source stream respectively. On the other hand, the target stream suffers from the absence of true class labels $B_{k_t}^T=\{x_{i}^T\}_{i=1}^{N_T}$ where $N_T$ is the size of the target stream. Note that we consider a case where both source and target domains are streaming in nature. $x_i^{S}\in\mathcal{X_S}$,    $x_i^{T}\in\mathcal{X_T}$, $\mathcal{X_S}\neq\mathcal{X_T}$ are input vectors of the source stream and the target stream while $y_i=[l_1,l_2,...,l_m]$ is a target vector formed as one-hot vector $y_i^S,y_i^T\in\mathcal{Y}$. $(x_i^S,y_i^S)\in\mathcal{X_S}\times\mathcal{Y}$ and $(x_i^T,y_i^T)\in\mathcal{X_T}\times\mathcal{Y}$. That is, the two domains feature different feature spaces but share the same labelling function, target variables, \textbf{Cross-Domain}. The two streaming data are sampled with different speeds resulting in $N_S\neq N_T$, \textbf{different batch sizes} while following different distributions $P(x_S)\neq P(x_T)$, \textbf{covariate shift}. The source stream and the target stream are non-stationary in nature where their concepts are drifting  $P(x,y)_t^S\neq P(x,y)_{t+1}^S$, $P(x,y)_{t'}^T\neq P(x,y)_{t'+1}^T$, $t\neq t'$, i.e., concept drifts of the two streams might develop at different time periods $t\neq t'$, \textbf{asynchronous drift}.

\section{Learning Procedure of LEOPARD}
     
    \subsection{Network Structure of LEOPARD}
	LEOPARD is structured as a deep clustering network developed with a feature extraction layer extracting natural features $Z$ from raw input features $x$ by means of a mapping function $F_{W_f}(.)$ where $W_f$ stands for parameters of the feature extractor. The extracted features $Z$ are passed to a fully connected layer formed as a stacked autoencoder (SAE) with a tied-weight constraint. That is, the decoder parameters are the inverse mapping of the encoder parameters. The natural features $Z\in\Re^{u'}$ are projected to a low dimensional latent space $h^l\in\Re^{R_l}$ where $u',R_l$ are respectively the number of natural features and hidden nodes at the $l-th$ layer $R_l<<u'$. The decoding and encoding mechanisms are expressed:
	\begin{equation}
	    h^{l}=r(W_{enc}^{l}h^{l-1}+b^l); h^0=Z
	\end{equation}
	\begin{equation}
	    \hat{h}^{l-1}=r(W_{dec}^{l}h^{l}+c^l);\quad \forall l = 1,\dots, L
	\end{equation}
	where $W_{enc}^l\in\Re^{R_l \times u_l},b^l\in\Re^{R_l}$ stand for the connective weights and biases of the $l-th$ layer of the encoder respectively while $W_{dec}^l\in\Re^{u_l\times R_l}, c^l\in\Re^{u_l}$ denote the connective weights and biases of the $l-th$ layer of the decoder respectively. The tied-weight constraint $W_{dec}^l=(W_{enc}^l)^T$ functions as a regularization mechanism preventing the issue of overfitting. 
	
\begin{figure}[!t]
	    \centerline{\includegraphics[scale=0.3]{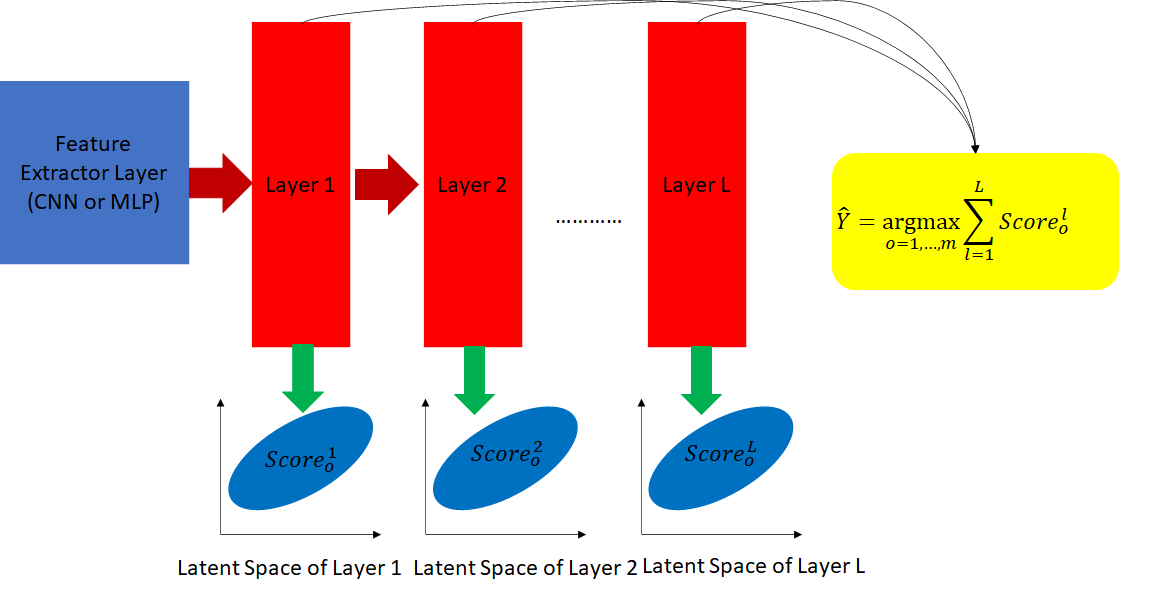}}
	    \caption{Network Structure of LEOPARD: LEOPARD adopts the different-depth network structure where the clustering module is implemented in every layer of SAE thus producing its own local outputs. The final predicted label is aggregated across different embedding layers.}
	    \label{fig:network structure}
	\end{figure}

	The clustering mechanism is carried out in each deep embedding space, each latent space. That is, it takes place in every hidden layer of SAE $h(.)^l$ creating different representations of data samples. The inference mechanism is performed by first calculating the similarity degree of a data sample and a hidden cluster\cite{DEC}:
	\begin{equation}\label{similarity}
	  \phi_{j}^{l}=\frac{(1+|h^l-C_{j}^l||_{2}/\lambda)^{\frac{-(\lambda+1)}{2}}}{\sum_{j=1}^{Clus^l}(1+|h^l-C_{j}^l||_{2}/\lambda)^{\frac{-(\lambda+1)}{2}}}
	\end{equation}
	where $C_j^l,h^l$ are the centroid of the $j-th$ cluster of the $l-th$ layer and the latent representation of a data sample $x$ of the $l-th$ layer while $Clus^l$ is the number of cluster created in the $l-th$ latent space, i.e., the $l-th$ layer of SAE. $\lambda=1$ is chosen here.  The student t-distribution is adopted to model the similarity degree and $\phi_j^l$ is also regarded as the cluster posterior probability $P(C_j^l|X)$ \cite{vandermaaten08a} where $P(C_j|X)=1$ presents the case of perfect match between $h^l$ and $C_j^l$. The similarity degree $\phi_{j}^l$ is aggregated across $N_m$ prerecorded samples having true class labels $B_0^S=\{x_i^S,y_i^S\}_{i=1}^{N_m}$. This operation produces the cluster's allegiance \cite{smith2019unsupervised} measuring cluster's tendencies to a particular class. Suppose that $N_o$ stands for the number of prerecorded samples having the $o-th$ class as their labels, the cluster allegiance $Ale_{j,o}^l$ is calculated:
	\begin{equation}\label{allegiance}
	    Ale_{j,o}^l=\frac{\sum_{n=1}^{N_o}\phi_{j,o}^{n,l}}{\sum_{o=1}^m\sum_{n=1}^{N_o}\phi_{j,o}^{n,l}}
	\end{equation}
	where $\phi_{j,o}^{n,l}$ measures the similarity degree of the cluster $C_j^l$ and the $n-th$ prerecorded sample $h_o^l$ falling into the $o-th$ class. (\ref{allegiance}) pinpoints the neighborhood degree of the $j-th$ cluster to the $o-th$ class and implies that an unclean cluster, occupied by data samples of mixed classes, possesses low cluster allegiance. The winner-takes-all principle $win=\arg\max_{j=1,...,Clus^l}\phi_{j}^l$ is adopted here, where a data sample is associated to the nearest cluster. The local score of the $l-th$ layer is defined as the allegiance of the winning cluster $Score^l=Ale_{win}^l$.
	 The predicted class label $\hat{Y}$ is determined as a class label maximizing its global score. The global score is calculated as the summation of a local score across $L$ layers: 
	\begin{equation}
	    \hat{Y}=\arg\max_{o=1,...,m}\sum_{l=1}^{L}Score^l
	\end{equation}
	where the majority voting approach is implemented here. It is evident that LEOPARD merely benefits from the labelled prerecorded samples of the source stream $B_0^S$ to associate a cluster to a specific class. No label at all from both streams is solicited in the streaming phase where it confirms its applicability in the extreme label scarcity environments. Fig. \ref{fig:network structure} visualizes the network structure of LEOPARD. It is perceived that the clustering process occurs in every hidden layer of LEOPARD thus producing its local outputs. The final predicted class label is aggregated across all layers making use of a summation operation. $R_l,L,Clus^l$ are self-evolved in respect to varying distributions. 
	
		\begin{figure*}[!t]
	    \centerline{\includegraphics[scale=0.45]{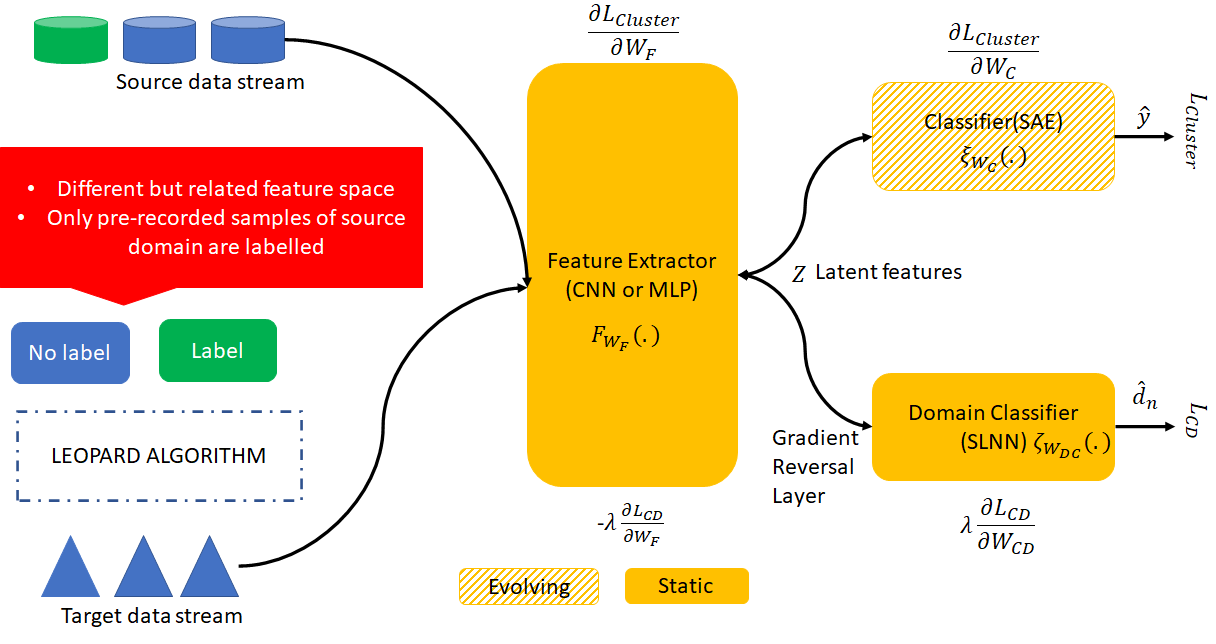}}
	    \caption{LEOPARD operates in the extreme label scarcity condition where only prerecorded samples of source stream are labelled while the rests are unlabelled. The learning algorithm of LEOPARD consists of three modules (feature extractor, classifier, domain classifier). Feature extractor generates latent features, classifier produces final prediction, domain classifier identifies sample origin. The feature extractor is updated by taking the gradient of the clustering loss and the cross domain loss. The gradient reversal layer is implemented to change the sign of the gradient of the cross domain loss. The classifier is updated by minimizing the clustering loss and the domain classifier is adjusted by minimizing the cross domain loss. The classifier features a self-evolving characteristic whereas the domain classifier and feature extractor are fixed.}
	    \label{fig:learning policy}
	\end{figure*}
	\subsection{Parameter Learning of LEOPARD}
	\noindent\textbf{Adversarial Domain Adaptation}: the idea of domain adaptation is to minimize the divergence between the target domain and the source domain. The concept of adversarial domain adaptation is founded by the idea of $H$ divergence \cite{DANN} where it relies on a hypothesis class $H$, a set of binary classifiers. 
	\textit{\textbf{Definition 1 \cite{Ben-david06}}}: Given the two domains $D_S$ and $D_T$ and the hypothesis class $H$, the $H$ divergence between $D_S$ and $D_T$ is defined as follows:
	\begin{equation}\label{H_divergence}
	\begin{split}
	    d_{H}(D_S,D_T) = 2\sup_{\eta\in H}|Pr[\eta(x)=1]_{x\sim D_S} \\ - Pr[\eta(x)=1]_{x\sim D_T}|
	    \end{split}
	\end{equation}
	The $H$ divergence in \eqref{H_divergence} relies on the hypothesis class $H$ to distinguish data samples generated from $D_S$ or data samples generated from $D_T$. In \cite{Ben-david06}, the empirical $H$ divergence can be used in the case of a symmetric hypothesis class $H$:
	\begin{equation}
	\begin{split}
	    d_{H}(D_S,D_T)=2(1-\min_{\eta\in H}[\frac{1}{n}\sum_{x\backsim D_S} I[\eta(x_n)=0]+\\\frac{1}{n^'}\sum_{x\backsim D_T}I[\eta(x_n)=1]])
	    \end{split}
	\end{equation}
	where $I[a]$ denotes an indicator function returning 1 if $a$ is true or 0 otherwise. This implies that the $H$ divergence can be minimized by finding a representation where the source and target samples are indistinguishable \cite{DANN}. 
	
	The concept of adversarial domain adaptation can be implemented by deploying a domain classifier $\zeta_{W_{DC}}(F_{W_F}(.))$ working along with the feature extractor $F_{W_f}(.)$ and a classifier $\xi_{W_{C}}(F_{W_F}(.))$. The domain classifier predicts the origin of data samples whether they are generated by the source domain $D_S$ or the target domain $D_T$ while the classifier generates the final output of a network. The domain reversal layer is implemented in updating the feature extractor such that indistinguishable features of source and target domains are induced. That is, the overall loss function is written as follows:
	\begin{equation}\label{adv_domain_adaptation}
	\begin{split}
	 L=\frac{1}{N_S}\sum_{n_1=1}^{N_S} L_{\xi}(\xi_{W_C}(F_{W_{f}}(x_{n_1}),y_{n_1})\\-\lambda(\frac{1}{N_S}\sum_{n_2=1}^{N_S}L_{\zeta}(\zeta_{W_{DC}}(F_{W_f}(x_{n_2}),d_{n_2})\\+\frac{1}{N_T}\sum_{n_3=1}^{N_T}L_{\zeta}(1-\zeta_{W_{DC}}(F_{W_f}(x_{n_3}),d_{n_3}))   
	\end{split}
	\end{equation}
	where $L_{\xi,\zeta}(.)$ is implemented as the cross entropy loss function and $d_n$ is the domain identity, i.e., 1 for the source domain and 0 for the target domain. From \eqref{adv_domain_adaptation}, the gradient reversal layer inserts a negative constant confusing the domain classifier, i.e., generating indistinguishable samples. The parameter learning process is formulated as follows:
	\begin{equation}\label{update1}
	    W_f=W_f-\mu(\frac{\partial L_{\xi}}{\partial W_{f}}-\alpha_1\frac{\partial L_{\zeta}}{\partial W_{f}})
	\end{equation}
	\begin{equation}\label{update2}
	    W_{C}=W_C-\mu\frac{\partial L_{\xi}}{\partial W_{C}}
	\end{equation}
	\begin{equation}\label{update3}
	    W_{DC}=W_{DC} - \mu\lambda\frac{\partial L_{\zeta}}{\partial W_{DC}}
	\end{equation}
	where the feature extractor is trained to produce similar features of the two domains seen in the negative sign of the gradient thereby leading to the domain invariant network. 
	\noindent\newline\textbf{Loss Function}: the parameter learning strategy of LEOPARD is constructed using a joint loss function comprising two modules: clustering loss and cross-domain adaptation loss. The underlying goal is to produce domain-invariant parameters as well as clustering-friendly latent spaces such that the online cross domain adaptation can be solved under extreme label scarcity. The overall cost function is formalized as follows:
	\begin{equation}\label{overall_loss}
	    L_{all}=L_{cluster}-\alpha_1 L_{cd}
	\end{equation}
	where $L_{cluster}, L_{cd}$ respectively denote the clustering loss and the cross-domain adaptation loss while $\alpha_1$ is a trade-off constant controlling the influence of the cross-domain adaptation loss. It is an unconstrained optimization problem which can be optimized using the stochastic gradient descent approach with no epoch or epoch per batch to assure scalability in streaming environments. That is, a number of iteration is done per batch. A data batch is discarded once iterations across a number of epoch is completed to allow bounded complexity. The negative sign in \eqref{overall_loss} follows the gradient reversal strategy generating similar features across two domains. In other words, the gradient of clustering loss and the gradient of cross domain loss, the domain classifier loss, is subtracted \cite{DANN}. 
	\noindent\newline\textbf{Clustering-Friendly Latent Space}: the clustering loss aims to achieve the clustering-friendly latent space via simultaneous feature learning and clustering. The clustering loss is formulated as the reconstruction loss and the KL divergence loss minimizing probabilistic distance of the latent space and the auxiliary target distribution \cite{DEC}:
	\begin{dmath}\label{clusloss}
	    L_{cluster}=\underbrace{L_{\xi}(x_{S,T},\hat{x}_{S,T})}_{L_1}+\underbrace{\sum_{l=1}^{L}(L_{\xi}(h_{S,T}^l,\hat{h}_{S,T}^l)\\+\alpha_2KL(\phi^l|\Phi^l))}_{L_2}
	\end{dmath}
	where $\Phi^l$ is the auxiliary target distribution of the $l-th$ latent space and $\alpha_2$ is a regularization constant controlling the strength of the KL divergence loss. $\phi^l$ is the similarity degree of the current sample to existing clusters. $L_{\xi}(.)$ is the reconstruction loss formed as the mean square error (MSE) loss function. It also performs nonlinear dimension reduction preventing the trivial solutions often happening in the case of linear mapping. It guarantees a data sample to be mapped back to its original representation. The key difference between the two loss functions lies in the adaptation mechanism in which $L_1$ is solved in the end-to-end fashion while $L_2$ is carried out in the layer-wise fashion. 
	
	The last term also known as the KL divergence loss minimizes the discrepancy of the distribution of a current data batch calculated via \eqref{similarity} and the auxiliary target distribution $KL(\phi^l|\Phi^l)=\sum_i\sum_j\phi_{i,j}^l\log{\frac{\phi_{i,j}^l}{\Phi_{i,j}^l}}$. The auxiliary target distribution should satisfy three requirements \cite{DEC}: 1) improve prediction; 2) emphasizes samples of high confidence; 3) normalize loss contribution of each cluster to avoid creations of large clusters. We adopt the same auxiliary distribution as in \cite{DEC} where $\Phi_{i,j}^l$ is raised to the second power and normalized by frequency per cluster:
	\begin{equation}
	    \Phi_{i,j}^l=\frac{(\phi_{i,j}^l)^2/\zeta_j}{\sum_{j=1}^{Clus^l}(\phi_{i,j}^l)^2/\zeta_j}
	\end{equation}
	where $\zeta_j=\sum_{i=1}^N\phi_{i,j}^l$ is the frequency of a cluster. This strategy is understood as the soft-cluster assignment \cite{DEC} where all clusters are updated and differs from the hard-cluster assignment only tuning the winning cluster. The clustering mechanism is hard to conduct in the high-dimensional space \cite{high_dimensional_stream_learning} thus calling for feature learning steps to be committed simultaneously. The clustering process takes place in every latent space $h(.)^l$ set as the common feature spaces between the source and target domain. That is,  \eqref{clusloss} is executed using samples of both source and target streams. This process also functions as an implicit domain adaptation strategy since the minimization of reconstruction loss across two streams with shared parameters ends up with an overlapping region of both domains \cite{ATL}. The optimization procedure takes place simultaneously where the network parameters and the cluster parameters are adjusted concurrently with the SGD method.
	\newline\noindent\textbf{Domain-Invariant Network}: LEOPARD consists of three sub-modules: feature extractor $F(.)$, classifier $\xi(.)$ and domain classifier $\zeta(.)$ to achieve a domain invariant property as depicted in Fig. \ref{fig:learning policy}. The feature extractor is parameterized by $W_{F}$ and the classifier formed as the deep clustering module is parameterized by $W_{C}\in\{W_{enc}^{l},W_{dec}^{l},C^{l}\}$ while the domain classifier formed as a single hidden layer network is parameterized by $W_{DC}$. The feature extractor and the domain classifier play a minimax game via the gradient reversal layer where the feature extractor is trained to fool the domain classifier via production of similar features of source and target streams while the domain classifier is trained to identify the origin of data samples. The cross domain adaptation loss is thus formulated as the domain classifier loss as follows:
	\begin{equation}
	\begin{split}
	    L_{cd} = \frac{1}{N_S}\sum_{n=1}^{N_S}L_{\zeta}(\zeta_{W_{DC}}(F_{W_f}(x_n)),d_n)\\+ \frac{1}{N_t}\sum_{n'=1}^{N_T}L_{\zeta}(1-\zeta_{W_{DC}}(F_{W_f}(x_{n'})),d_{n'})
	 \end{split}   
	\end{equation}
	where $d_n$ stands for the origin of data samples, i.e., $1$ for the source stream and $0$ for the target stream. The domain classifier is tasked to solve a binary classification problem where $L_{\zeta}(.)$ is set as the cross entropy loss function. This leads to similar parameter learning processes for feature extractor, domain classifier and classifier respectively as in \eqref{update1} - \eqref{update3} except the presence of the clustering loss instead of the cross-entropy loss as defined in \eqref{clusloss}: $W_{f}=W_{f}-\mu(\frac{\partial L_{cluster}}{\partial W_{f}}-\alpha_1\frac{\partial L_{cd}}{\partial W_{f}});W_{C}=W_{C}-\mu\frac{\partial L_{cluster}}{\partial W_{C}}; W_{DC}=W_{DC}-\mu\alpha_1\frac{\partial L_{cd}}{\partial W_{DC}}$ where $\mu$ denotes the learning rate. Note that the gradient reversal layer has no parameters and simply alters the sign of the gradients allowing maximization process to be carried out via the stochastic gradient descent approach. This only applies to the feature extractor as illustrated in Fig. \ref{fig:learning policy}.  
	 
	\subsection{Structural Learning of LEOPARD}
    \noindent\textbf{Evolution of Cluster}:  The classifier of LEOPARD implements the self-organizing mechanism of network clusters where the clusters are flexibly grown in every hidden layer $h(.)^l$ if changing data distributions are identified. Furthermore, it is performed for both source data samples $h_S^l$ and target data samples $h_T^l$. That is, the clustering mechanism does not generate stream-specific clusters. Suppose that $D(X,Y)$ stands for the $L_2$ distance between two variables $X,Y$ and an $i-th$ cluster of $l-th$ layer is parameterized by its centre $C_i^l$, the growing condition is formulated as follows:
    \begin{equation}\label{clusadd}
        \min_{i=1,...,Clus^l}D(h^l,C_{i}^l)>\mu_{D,i}^l+k_1\sigma_{D,i}^l
    \end{equation}
    where $\mu_{D,i}^l,\sigma_{D,i}^l$ denote the mean and standard deviation of the distance $D(h^l,C_i^l)$ of the $i-th$ cluster of the $l-th$ layer while $k_1=2\exp{-||h_l-C_{win}^l||}+2$ leading to a dynamic confidence degree. The dynamic confidence degree enables the cluster growing phase to be carried out in the case of a far proximity between a data sample and the winning cluster. (\ref{clusadd}) examines the coverage span of existing clusters where a new cluster is inserted if a data sample is remote from the influence zone of existing clusters or a concept drift develops. A new cluster is crafted by assigning the current sample of interest as the cluster's center $C_{Clus^l+1}^l=h^l$ and setting the cluster's cardinality to be $N_{Clus^l+1}=1$. 
    \noindent\newline\textbf{Evolution of Network Structure}: The classifier of LEOPARD is equipped by the hidden node growing and pruning strategies adapting to the concept drifts of data streams. That is, this mechanism takes place for both the source stream and the target stream. The self-organizing mechanism is controlled by the network significance (NS) method \cite{nadine} adopting the bias-variance decomposition concept of every layer. That is, a high bias situation leads to an introduction of a new node while a high variance condition triggers the node pruning mechanism. Note that the network bias and variance here are evaluated in respect to the local error of a layer. All of which are carried out in an unsupervised fashion in respect to the reconstruction error. The network significance (NS) method is formalized as follows:
    \begin{equation}\label{NS}
        NS=(E[\hat{h}^l]-h^l)^2+(E[(\hat{h}^l)^2]-E[\hat{h}^l]^2)
    \end{equation}
    (\ref{NS}) can be solved by finding the expected output $E[\hat{h}^l]$ under a certain probability density function $p(x)$ assumed to follow the normal distribution $N(\mu,\sigma^2)$ with mean $\mu$ and variance $\sigma^2$. The bottleneck of this approach is found in the case of drift $p(x)_t\neq p(x)_{t+1}$ where it does not keep pace with rapidly changing distributions. To correct this shortcoming, Autonomous Gaussian Mixture Model (AGMM) can be used to estimate a complex probability density function $p(x)$ as done in \cite{ATL}. It is computationally expensive and often unstable in the high input dimension case due to the use of product norm. Furthermore, we deal with a multi-layer network here doubling the complexity of AGMM. 
    
    The hidden unit growing and pruning steps are signalled by the statistical process control (SPC) approach \cite{GamaDataStream} commonly used for anomaly detection tasks. The SPC method is applied here to detect the high bias or high variance condition and written as follows:
    \begin{equation}\label{growing}
	    \mu_{bias}^{n,l}+\sigma_{bias}^{n,l}\geq\mu_{bias}^{min,l}+k_2\sigma_{bias}^{min,l}
	\end{equation}
	\begin{equation}\label{pruning}
	    \mu_{var}^{n,l}+\sigma_{var}^{n,l}\geq\mu_{var}^{min,l}+2*k_3\sigma_{var}^{min,l}
	\end{equation}
	The SPC method is generalized here using $k_2=1.3\exp{(-Bias^2)}+0.7$ and $k_3=1.3\exp{(-Var^2)}+0.7$. This modification leads to dynamic confidence levels enabling for flexible growing and pruning phases. That is, the node growing process is likely performed in the case of a high bias while being strict in the case of a low bias. The same case also applies for the node pruning mechanism. $\mu_{bias}^{min,l}$ and $\sigma_{bias}^{min,l}$ are reset if the growing condition (\ref{growing}) is satisfied. On the other hand, if the pruning condition is met, $\mu_{var}^{min,l}$ and $\sigma_{var}^{min,l}$ are reset. The initialization of a new node is carried out using the Xavier initialization strategy. The least contributing node having the least statistical contribution is subject to the pruning step if (\ref{pruning}) is observed. Since LEOPARD is constructed under a different-depth structure where every layer performs its own clustering mechanism and produces its local output, the growing and pruning steps are independently undertaken per layer. Furthermore, this mechanism occurs for both source and target streams to anticipate the asynchronous drift problem where the network structure is shared across two domains. 
	
	The classifier of LEOPARD is fitted with the hidden layer growing mechanism where it expands the network depth based on the drift detection mechanism \cite{DEVFNN}. The drift detection mechanism is designed from the concept of Hoeffding's bound and analyzes the dynamic of latent features $Z$ to identify the change of marginal distribution. Note that no labelled samples are offered for model updates and the drift detection approach is executed for both source and target streams. The addition of a network layer is desired in practise because it is capable of substantiating network capacity significantly thus enhancing model's generalization. The drift detection procedure starts by finding the cutting point, a point where population mean increases. A cutting point is declared by the following condition.
    \begin{equation}
    \hat{P}+\epsilon_P\geq\hat{Q}+\epsilon_Q    
    \end{equation}
    where $P\in\Re^{2N}$ is a data matrix containing two consecutive data batches $[B_{k-1},B_k]$, i.e., previous and current data batches while $Q\in\Re^{cut}$ is a data matrix with $cut$ as the hypothetical cutting point of interest, $cut<2N$. Two data batches are applied here to increase the sensitivity of cutting point identification because latent features are relatively stable compared to the original input space. The hypothetical cutting point is arranged as $cut=[25\%,50\%,75\%]\times 2N$ instead of every point to avoid false alarm. $\hat{P},\hat{Q}$ denote the statistics of data matrices $P,Q$. $\epsilon_{P,Q}$ stand for the error bound derived from the concept of Hoeffding's bound as follows:
    \begin{equation}
        \epsilon_{P,Q}=\sqrt{\frac{1}{2\times size}\ln{\frac{1}{\alpha_x}}}
    \end{equation}
	where $\alpha_x$ is the significance level being inversely proportional to the confidence level $1-\alpha_x$ while $size$ refers to the size of the data matrix of interest $P,Q$. 
	
	Once eliciting the cutting point of interest $cut$, a data matrix $R\in\Re^{2N-cut}$ is constructed. A drift is signalled if $|\hat{R}-\hat{Q}|\geq\epsilon_{D}$. Beside the drift condition, a warning condition is set and pinpoints a case where a drift needs to be confirmed by the next data batch. That is, $\epsilon_{W}\leq|\hat{R}-\hat{Q}|\leq\epsilon_{D}$ where $\alpha_{W}<\alpha_{D}$. The error bounds $\epsilon_{D,W}$ are defined as follows:
	\begin{equation}\label{drift}
	    \epsilon_{D,W}=(b-a)\times\sqrt{\frac{size-cut}{2\times cut\times size}\ln{\frac{1}{\alpha_{D.W}}}}
	\end{equation}
	where $[a,b]$ denotes the range of the data matrix $P$. A new layer is created if a concept drift is found. That is, the number of nodes is set as the half of the network width of the previous layer $l-1$. This step enables the nonlinear feature reduction and avoids an over-complete network. The domain classifier and the feature extractor have a fixed structure because the structural learning of the classifier suffices to address the asynchronous drift problem. 
	
	\subsection{Algorithm}
	Learning policy of LEOPARD is visualized in Fig. \ref{fig:learning policy} and Algorithm 1 where LEOPARD is driven by the feature extractor, the classifier and the domain classifier. The forward pass procedure is done by feeding raw input attributes $x_{S,T}$ to the feature extractor $F(.)$ leading to latent input features $Z_{S,T}$. The latent features are passed to the classifier $\xi(.)$ implemented as the SAE and the clustering module. Note that the clustering module exists in every layer of SAE producing its own local output $Score^l$ where the majority voting is performed to generate a final predicted output. The learning process starts with a warm-up phase using $N_{init}$ unlabelled samples iterated across $E$ number of epochs to avoid the cold start problem. This process only involves the reconstruction loss $L_{\xi}(x_{S,T},\hat{x}_{S,T})$ and $L_{\xi}(h^{l}_{S,T},\hat{h}^{l}_{S,T})$ affecting only network parameters $W_{F}$ and $W_{enc}^l, W_{dec}^l$. The main training loop is executed by minimizing $L_{cluster}(.)$ and is applied to $W_{F},W_{enc}^{l},W_{dec}^{l},C^{l}$. Minimization of clustering loss across the two domains can be also seen as the domain adaptation strategy because it leads to an overlapping region of source domain and target domain to be created, i.e., both the source stream and the target stream are used under shared parameters. The adversarial domain adaptation is carried out by minimizing $L_{cd}(.)$ afterward where the domain classifier $\zeta_{W_{DC}}(.)$ is updated as well as the feature extractor $F_{W_{f}}(.)$ using the cross domain loss. The gradient reversal strategy is adopted when adjusting the feature extractor thus converting the minimization problem to the maximization problem and in turn resulting in indistinguishable features of the source stream and the target stream. The cross domain adaptation strategy makes possible for the source streams and the target streams following different distributions to be mapped similarly, i.e., the covariate shift is addressed. 
	
	\begin{algorithm}
      \caption{LEOPARD}\label{alg:leopard algorithm}
      \KwIn{Source streaming data \{$B_0^S, B_1^S, B_2^S, ..., B_{K_S}^S$\}, target streaming data \{$B_1^T, B_2^T, ..., B_{K_T}^T$\}, initialization epochs $E_{init}$, batch number of source and target streaming data $b_k$, epoch number $E$.}
      \KwOut{Network parameters of feature extractor $W_f$, classifier $W_C$ and domain classifier $W_{DC}$. Average accuracy $Acc$.}
      \For{$i=1:E_{init}$}
          {
            Initializing clusters using scarcity labelled data $B_0^S$\;
          }
      \For{$j=1:E$}
      {
        Network layer evolution of classifier (SAE) $\xi_{W_c}$ by Eq. (\ref{drift})\;
        Hidden unit of classifier (SAE) $\xi_{W_c}$ growing and pruning by Eq. (\ref{growing}) and (\ref{pruning})\;
         {$L_{cluster}={L_{\xi}(x_{S,T},\hat{x}_{S,T})}+{\sum_{l=1}^{L}(L_{\xi}(h_{S,T}^l,\hat{h}_{S,T}^l)+\alpha_2KL(\phi^l|\Phi^l))}$\;}
         Update feature extractor parameter $W_f$ and classifier parameter $W_C$ in respect to $L_{cluster}$\;
         {$L_{cd}= \frac{1}{N_S}\sum_{n=1}^{N_S}L_{\zeta}(\zeta_{W_{DC}}(F_{W_f}(x_n)),d_n)+ \frac{1}{N_t}\sum_{n'=1}^{N_T}L_{\zeta}(1-\zeta_{W_{DC}}(F_{W_f}(x_{n'})),d_{n'})$}\;
         Update feature extractor parameter $W_f$ and domain classifier parameter $W_{DC}$ in respect to $L_{cd}$\;
         }
      return $W_f, W_C, W_{DC}$ and average accuracy $Acc$\;
    \end{algorithm}

	The structural learning process occurs in both the initialization phase and the main training phase in which it includes the cluster growing process, the hidden node growing and pruning processes and the hidden layer growing process. As with the warm-up phase, the initialization phase using $N_{init}$ prerecorded samples over $E$ epochs is implemented if a new layer is created. It is obvious that LEOPARD does not exploit any labelled samples for model updates except for labelled samples to be used to calculate the cluster allegiance \eqref{allegiance}. The structural learning mechanism addresses the issue of asynchronous drifts across both streams.  
    \section{Numerical Study}
    This section presents numerical validation of LEOPARD putting forward nine datasets leading to 24 independent numerical results. Ablation study is added in this section to further numerically validate the contribution of each learning component. Source codes of LEOPARD can be found in \url{https://github.com/wengweng001/LEOPARD.git}. Our analysis of label proportions and visualizations of LEOPARD's learning performances are offered in the supplemental document. 
    \subsection{Dataset}
    \noindent\textbf{MNIST(MN)}$\leftrightarrow$\textbf{USPS(US)}:
    this problem presents a digit recognition problem having 10 classes. The data samples are formed by gray-scale images of hand-written digits resized to $28\times28$ for US$\rightarrow$MN and $28\times28$ for MN$\rightarrow$US cases. 
    \newline\noindent\textbf{Amazon@X(AM)}: this is a multi-domain sentiment analysis problem encompassing product reviews obtained from Amazon.com. $X$ stands for the product type \cite{AmazonReview}. Five product types, namely beauty, books, industrial, luxury and magazine, are selected here where the cross-domain multistream classification problem is formulated with two products with similar contexts but different topics. The averaged summed outputs from Google's word2vec model pretrained on 100 billion words \cite{word2vec} is used to perform feature extraction.  
    \newline\noindent\textbf{Office31}: this problem presents three domains: amazon (A), DSLR (D) and Webcam (W). It comprises 31 categories of the office objects. We present the case of D$\leftrightarrow$W where D comprises 498 images and W consists of 795 images. The characteristics of nine datasets are summed up in Table \ref{tab:dataset_char}.
    \subsection{Simulation Protocol}
    The numerical study is carried out using the prequential test-then-train protocol as per \cite{GamaDataStream}, A model is tested first before updating it with the same data stream. 
    The numerical evaluation is independently undertaken per-batch where the numerical result is averaged across all batches. Our simulation is repeated 5 times to guarantee the consistency of numerical results where the final numerical results are averaged over 5 independent runs. The asynchronous drift problem is induced by applying the scaling hyper-plane strategy \cite{Gamaconceptdrift,ACDC} where a data stream is scaled to $x_i=\frac{d_z\times x_i}{||x||}$. $d_z$ is a randomly generated concept drift vector where $z$ is the number of concept drifts in the stream: $z=1$ for every source stream and $z=1$ for every target data stream. A fixed random seed is selected in setting $d_z$ to assure fair comparison. In realm of MN$\leftrightarrow$US, the concept drifts occurs at $k=35$ for source stream and $k=36$ for target stream whereas the concept drift takes place at $k=5$ for source stream and $k=6$ for target stream in the amazon@X and Office 31 problems. These configurations assure the asynchronous drift to be presented. 
    
    \begin{table}[]
        \centering
         \caption{Characteristics of Datasets}
        \begin{tabular}{c|c|c|c|c}
        \toprule
            Dataset & Attributes & Labels & Samples & NB\\
        \midrule    
          MNIST(MN)       & 784   & 10    & 70000 & 65 \\
          USPS(US)    & 256   & 10    & 9298  & 65\\
          Amazon@Beauty(AM1) & 300   & 5 & 5150  & 20\\
          Amazon@Books(AM2)  & 300   & 5 & 500000 & 20   \\
          Amazon@Industrial(AM3) & 300   & 5 & 73146 & 20\\
          Amazon@Luxury(AM4) & 300   & 5 & 33784 & 20\\
          Amazon@Magazine(AM5)   & 300   & 5 & 2230 & 20 \\
          Office31(D) & 36636672 & 31& 498 & 10 \\
          Office31(W) & 921600 & 31 & 795 & 10 \\
          \bottomrule
        \end{tabular}
        NB: Number of Batches
      \label{tab:dataset_char}
    \end{table}
    
    \subsection{Baseline}
    LEOPARD is compared with five algorithms: autonomous deep clustering network (ADCN) \cite{Ashfahani2022UnsupervisedCL}, deep clustering network (DCN) \cite{clustering_friendly}, autoencoder followed by K-Means (AE+KMeans), deep embedding clustering (DEC) \cite{DEC} and domain adversarial neural networks (DANN) \cite{DANN}. ADCN is a self-evolving deep clustering network where hidden clusters, nodes and layers are grown and pruned dynamically. The loss function is formulated with a combination of a clustering loss and a reconstruction loss. ADCN is not equipped by a specific domain adaptation loss function while it applies the hard cluster assignment approach as with \cite{clustering_friendly}, i.e., The $L_2$ distance loss of the winning cluster and the latent sample is put forward. DCN adopts a fixed network structure where the clustering mechanism only takes place at the bottleneck layer. It applies the same loss function as ADCN. AE+KMeans differs from DCN where the clustering mechanism is carried out after the training process. It does not utilize any clustering loss. DEC adopts the soft-assignment approach as with LEOPARD except that it relies on a static network structure and suffers from the absence of any domain adaptation loss. The reconstruction loss in the baseline algorithms are perceived as a domain adaptation procedure because they are carried out for both source and target streams under shared network parameters. DANN utilizes the adversarial domain adaptation as per LEOPARD without any clustering mechanism.
    
    All of them work under the extreme label scarcity condition as with LEOPARD where access of true class labels is only provided for the prerecorded samples of the source stream while no label is offered during the process runs for both the source stream and the target stream. Comparison with ADCN is done by executing their published codes to assure fair comparisons. We utilize our own implementations of DCN, AE+KMeans, DEC and DANN.
    \subsection{Hyperparameters}
    The learning rate and momentum of LEOPARD are allocated as 0.01 and 0.95 while the regularization constant of the clustering loss $\alpha_2$ is set as 1 and the tradeoff constant of the cross-domain loss $\alpha_1$ is set as 0.1. LEOPARD also depends on labelled prerecorded samples $B_0^S$ of the source stream set as $10\%$ of source samples proportionally taken from each class $N_m=10\%N_S$. That is, each class contributes the same number of samples. The number of initial epochs are set as $E=100$ (amazon@X), $E=50$ (MN$\leftrightarrow$US), and $E=500$ (D$\leftrightarrow$W) respectively. The initialization phase is carried out using labelled prerecorded samples of the source stream. The parameters of the drift detector $\alpha_x,\alpha_D,\alpha_W$ are selected respectively as $0.001,0.001,0.005$. For amazon@X problems, LEOPARD runs in the one-pass learning procedure whereas for MN$\leftrightarrow$US experiments, the training process of the clustering loss adopts the epoch per batch strategy with $10$ (MN$\rightarrow$US, W$\leftrightarrow$D) epochs and $5$ epochs (US$\rightarrow$MN) respectively. The epoch per batch strategy satisfy the online learning requirement because a data batch is discarded after training over predetermined epochs. The same setting is also applied to the baseline algorithms assuring fair comparisons.
    
    For MN$\leftrightarrow$US problem, the feature extractor is formed as convolutional neural networks. The encoder part is constructed as 2 convolutional layers using 16 and 4 filters respectively while having the max pooling layer in between. The decoder part is built upon two transposed convolutional layers with 4 and 16 filters respectively. For amazon@X sentiment analysis problems, the multi-layer perceptron feature extractor is put forward with two hidden layers where the number of nodes is fixed as $300$ and $100$. For the office31 problem, ResNet34 is applied as feature extractors. The initial nodes of fully connected layer are simply assigned as $96$ for the MNIST$\leftrightarrow$USPS problem, $30$ for amazon@X sentiment analysis problems and $500$ for D$\leftrightarrow$W. The ReLU activation function is applied for the intermediate layers while the decoder output utilizes the sigmoid activation function producing normalized reconstructed output. The network structures of baseline algorithms are set similarly to ensure fair comparison. Further details of our numerical studies are explained in the LEOPARD's codes shared in \url{https://github.com/wengweng001/LEOPARD.git}
    
    These parameters are fixed throughout all study cases to guarantee non ad-hoc performance of LEOPARD.  The hyper-parameters of the baselines are selected as per the guidelines of their publications and hand-tuned if their performances are surprisingly compromised. The hyper-parameters of all consolidated algorithms are listed in the supplemental document.

\begin{table*}[htbp]
  \centering
  \caption{Average Accuracy (\%) of The Target Stream across 5 runs,   *indicates statistically significant results and \textbf{BOLD} denotes the best numerical results}
 \begin{threeparttable}
    \begin{tabular}{ccccccc}
    \toprule
    Experiments & LEOPARD & ADCN  & AE-kmeans & DCN   & DEC & DANN\\
    \midrule
    AM1 $\rightarrow$ AM2 & 20.6320 $\pm$ 2.3958 & 19.8160 $\pm$ 4.7555 & \textbf{27.8012 $\pm$ 1.6392} & 27.7792 $\pm$ 1.8319 & 18.3774 $\pm$ 1.3356 & 15.2291 $\pm$ 22.7966\\
    AM1 $\rightarrow$ AM3 & \textbf{*71.5300 $\pm$ 1.0819} & 57.0520 $\pm$ 10.7930 & 25.8713 $\pm$ 1.6412 & 26.1010 $\pm$ 1.9047 & 17.2324 $\pm$ 1.5599 & 34.8559 $\pm$ 29.7157\\
    AM1 $\rightarrow$ AM4 & \textbf{*57.7840 $\pm$ 0.0476} & 43.9800 $\pm$ 3.3004 & 27.9307 $\pm$ 1.2602 & 28.1326 $\pm$ 1.1633 & 16.6092 $\pm$ 0.6464 & 44.4809 $\pm$ 20.6428\\
    AM1 $\rightarrow$ AM5 & \textbf{*63.5100 $\pm$ 1.2016} & 60.7980 $\pm$ 3.0386 & 31.3240 $\pm$ 0.8684 & 31.1996 $\pm$ 0.8381 & 13.9947  $\pm$ 1.0473 & 41.5806 $\pm$ 13.6181\\
    AM2 $\rightarrow$ AM1 & \textbf{*71.5480  $\pm$ 8.8031} & 25.2880  $\pm$ 6.6382 & 36.7868  $\pm$ 1.5799 & 36.9540  $\pm$ 2.0093 & 8.8334  $\pm$ 0.9026 & 49.4693 $\pm$ 39.4318\\
    AM2 $\rightarrow$ AM3 & \textbf{45.4920  $\pm$ 13.2055} & 19.9380  $\pm$ 15.1533 & 31.0612  $\pm$ 1.8636 & 30.9986  $\pm$ 1.8202 & 14.2852  $\pm$ 1.0375 & 43.4064 $\pm$ 28.3212\\
    AM2 $\rightarrow$ AM4 & \textbf{*48.5600  $\pm$ 3.5658} & 14.1460  $\pm$ 1.7088 & 27.1297  $\pm$ 1.0116 & 27.2251  $\pm$ 0.9788 & 15.6150  $\pm$ 1.4221 & 42.8489 $\pm$ 14.2926\\
    AM2 $\rightarrow$ AM5 & 50.3680  $\pm$ 15.8943 & 31.9160  $\pm$ 5.8286 & 28.7212  $\pm$ 1.6700 & 28.4330  $\pm$ 1.0860 & 18.3333  $\pm$ 2.6212 & \textbf{60.4227 $\pm$ 8.1435}\\
    AM3 $\rightarrow$ AM1 & 37.3520  $\pm$ 4.7246 & \textbf{53.0180  $\pm$ 17.7237} & 25.7504  $\pm$ 1.2193 & 25.4591  $\pm$ 1.4807 & 7.7442  $\pm$ 1.5116 & 17.1871 $\pm$ 19.2744\\
    AM3 $\rightarrow$ AM2 & 31.3120  $\pm$ 8.9392 & 8.9900  $\pm$ 2.0600 & 22.1118  $\pm$ 1.1815 & 25.3291  $\pm$ 5.8252 & 17.2165  $\pm$ 1.6547 & \textbf{40.9799 $\pm$ 13.1832}\\
    AM3 $\rightarrow$ AM4 & 18.7240  $\pm$ 1.0962 & \textbf{28.4340  $\pm$ 4.5292} & 23.2826  $\pm$ 1.4628 & 23.1707  $\pm$ 1.6232 & 15.2796  $\pm$ 2.2464 & 22.2822 $\pm$ 17.1165\\
    AM3 $\rightarrow$ AM5 & \textbf{*59.5520  $\pm$ 2.8339} & 37.1540  $\pm$ 8.1052 & 22.1968  $\pm$ 0.9799 & 22.3941  $\pm$ 0.6026 & 16.0303  $\pm$ 2.0662 & 20.1265 $\pm$ 21.2838\\
    AM4 $\rightarrow$ AM1 & 45.5560  $\pm$ 6.3118 & \textbf{69.4040 $\pm$ 6.0339} & 23.2919 $\pm$ 3.1591 & 23.4620 $\pm$ 3.0392 & 8.0682 $\pm$ 0.7888 & 55.4146 $\pm$ 40.1126\\
    AM4 $\rightarrow$ AM2 & \textbf{23.2340 $\pm$ 5.5658} & 21.9140 $\pm$ 6.7656 & 21.5370 $\pm$ 0.8304 & 21.3970 $\pm$ 0.9446 & 18.1515 $\pm$ 1.2285 & 22.8453 $\pm$ 19.9388\\
    AM4 $\rightarrow$ AM3 & 58.0500 $\pm$ 4.2461 & \textbf{62.6160 $\pm$ 3.7864} & 22.4032 $\pm$ 1.3615 & 22.6766 $\pm$ 1.4520 & 17.2892 $\pm$ 1.9414 & 54.9784 $\pm$ 22.9740\\
    AM4 $\rightarrow$ AM5 & \textbf{*64.3480 $\pm$ 0.0895} & 56.0280 $\pm$ 2.6346 & 21.3376 $\pm$ 1.1374 & 21.3101 $\pm$ 0.9658 & 15.5601 $\pm$ 1.4566 & 20.0455 $\pm$ 13.7290\\
    AM5 $\rightarrow$ AM1 & \textbf{*87.6760 $\pm$ 0.3844} & 56.3760 $\pm$ 19.8715 & 20.1306 $\pm$ 1.2201 & 20.4453 $\pm$ 0.8831 & 8.8645 $\pm$ 1.4946 & 61.0045 $\pm$ 17.2396\\
    AM5 $\rightarrow$ AM2 & 12.5060 $\pm$ 1.1611 & 10.8880 $\pm$ 3.3736 & 19.3033 $\pm$ 1.0140 & \textbf{19.6829 $\pm$ 0.8429} & 15.7549 $\pm$ 2.1423 & 36.7490 $\pm$ 25.1993\\
    AM5 $\rightarrow$ AM3 & \textbf{*36.7900 $\pm$ 6.2853} & 27.9480 $\pm$ 3.3931 & 21.4889 $\pm$ 2.2889 & 21.1898 $\pm$ 2.2209 & 16.9886 $\pm$ 3.9668 & 31.0573 $\pm$ 32.7811\\
    AM5 $\rightarrow$ AM4 & \textbf{*50.7580 $\pm$ 2.7020} & 32.5880 $\pm$ 4.3853 & 19.8205 $\pm$ 0.7625 & 19.5470 $\pm$ 0.6045 & 16.0969 $\pm$ 1.2545 & 33.5787 $\pm$ 21.3489\\
    \midrule
    MNIST $\rightarrow$ USPS & 45.3740 $\pm$ 14.3497 & \textbf{62.9800 $\pm$ 3.0666}     & 10.1138 $\pm$ 0.2647 & 9.9913 $\pm$ 0.3020 & 10.3657 $\pm$ 1.5318 & 23.1336 $\pm$ 3.0172\\
    USPS $\rightarrow$ MNIST & \textbf{*49.4660 $\pm$ 2.1841} & 33.3800 $\pm$ 14.5818 & 10.0563 $\pm$ 0.3269 & 9.6323 $\pm$ 0.8033 & 9.3434 $\pm$ 0.5433 & 39.0328 $\pm$ 6.7573\\
    \midrule
    D $\rightarrow$ W & \textbf{*41.8080 $\pm$ 9.8034} & 4.0000 $\pm$ 0.5589 & 3.7722 $\pm$ 0.3789 & 3.0633 $\pm$ 0.7778 & 3.2658 $\pm$ 0.6425 & 10.9821 $\pm$ 2.7133\\
    W $\rightarrow$ D & \textbf{*35.2820 $\pm$ 12.0613} & 3.7560 $\pm$ 0.4568 & 2.9388 $\pm$ 0.6657 & 3.1429 $\pm$ 0.7370 & 2.8163 $\pm$ 0.4725 & 6.4323 $\pm$ 1.8659 \\
    \bottomrule
        \end{tabular}%
        \begin{tablenotes}
              \footnotesize
              \item[*] The standard for statistically significant results is based on the T-Test score between LEOPARD and other baselines, 4 T scores for each experiment. We determined the result as statistically significant if t score is greater than 2.015 for at least 3 out of 4.
        \end{tablenotes}

        \end{threeparttable}
      \label{tab:addlabel}%
    \end{table*}%


    \subsection{Numerical Results}
    From Table \ref{tab:addlabel}, it is seen that LEOPARD outperforms other algorithms in 15 of 24 cases with noticeable margins. This aspect portrays the efficacy of the adversarial domain adaptation approach and the soft-cluster assignment mechanism where these two modules are absent in the baseline algorithms, i.e., ADCN, DCN, AE+KMeans adopt the hard-cluster assignment strategy and suffer from the absence of the adversarial domain adaptation approach. This finding also confirms the advantage of the adversarial domain adaptation over the feature reconstruction strategy with shared parameters across the two streams implemented in all baselines. The soft-cluster assignment approach where the cluster and network parameters are simultaneously optimized via the SGD method performs better than the hard cluster assignment strategy. It is demonstrated by the fact that LEOPARD beats ADCN in significant numbers of cases. There is no significant performance difference between AE+KMEANS and DEC. On the other hand, the importance of the structural learning component in handling data streams is clearly portrayed here where LEOPARD and ADCN are superior compared to other algorithms having static structures. Such mechanism allows timely reactions to the asynchronous drift problem across the source stream and the target stream. The performance of DANN implemented under conventional neural network structures is far inferior to LEOPARD under extreme label scarcity condition. This fact confirms the advantage of clustering approach compared to the conventional neural network structure to reduce label's dependencies. The statistical test is undertaken using the t test $(P<0.05)$ confirming the advantage of LEOPARD where it beats other algorithms with statistically significant gaps in 13 of 24 cases. 
    
    \begin{table*}[htbp]
      \centering
      \caption{Ablation Study of LEOPARD}
      \begin{threeparttable}
        \begin{tabular}{cccccc}
        \toprule
              & A     & B     & C     & D  & E \\
        \midrule
        AM1 $\rightarrow$ AM3 & 56.6620 $\pm$ 7.3010 & 70.0860 $\pm$ 2.3159 & 51.7800 $\pm$ 5.3515 & 40.8860 $\pm$ 2.6187 & 71.5300 $\pm$ 1.0819 \\
        AM1 $\rightarrow$ AM4 & 49.4800 $\pm$ 3.9323 & 57.2160 $\pm$ 0.3800 & 49.3940 $\pm$ 3.5215 & 27.7520 $\pm$ 1.9921 & 57.7840 $\pm$ 0.0476\\
        AM5 $\rightarrow$ AM1 & 85.6580 $\pm$ 2.3485 & 83.2040 $\pm$ 6.6126 & 86.7800 $\pm$ 1.5401 & 31.1860 $\pm$ 1.5308 & 87.6760 $\pm$ 0.3844\\
        AM5 $\rightarrow$ AM4 & 29.6480 $\pm$ 5.5269 & 43.5760 $\pm$ 6.5908 & 29.4860 $\pm$ 3.4137 & 24.7100 $\pm$ 1.6597 & 50.7580 $\pm$ 2.7020 \\
            \bottomrule
        \end{tabular}%
        \begin{tablenotes}
              \footnotesize
              \item[A] LEOPARD model without network structure evolution and additional loss ($KL(\phi^l|\Phi^l)$ and $L_{cd}$)
              \item[B] LEOPARD model without network structure evolution
              \item[C] LEOPARD model with the absence of $KL(\phi^l|\Phi^l)$ and $L_{cd}$
              \item[D] LEOPARD model using BERT as the feature extractor
              \item[E] LEOPARD model
        \end{tablenotes}

        \end{threeparttable}
      \label{ablationstudy}%
    \end{table*}%
    
    \subsection{Ablation Study}
    This section studies the effect of LEOPARD's learning modules by analyzing its performance when deactivating a particular learning module. LEOPARD is configured into four models: (A) the structural learning method is switched off leaving LEOPARD to have a static network structure. In addition, the parameter learning strategy is done with the absence of the cross domain adaptation loss and the KL divergence loss. In short, LEOPARD is driven by the reconstruction loss only; (B) the structural learning strategy of LEOPARD is deactivated while the parameter learning step utilizes both the cross domain adaptation loss and the KL divergence loss; (C) the structural learning mechanism of LEOPARD is activated but with the absence of the KL divergence loss and the cross domain adaptation loss; (D) BERT is applied as a feature extractor in lieu of a word2vec model. Our ablation study is carried out using four study cases: AM1$\rightarrow$AM3, AM1$\rightarrow$AM4, AM5$\rightarrow$AM1 and AM5$\rightarrow$AM4.
    
    From table \ref{ablationstudy}, LEOPARD suffers from major performance degradation for all configurations (A)-(D) confirming the efficacy of its current version. For AM5$\rightarrow$AM1, configuration (A) where both the structural learning strategy, the KL divergence loss and the cross domain adaptation loss are absent produces poor results. The performance improves slightly with the activation of the KL divergence loss and the cross domain adaptation loss as per configuration (B). Although the KL divergence loss and the cross domain adaptation loss are deactivated, the structural learning mechanism improves the accuracy as per configuration (C) but is not yet on par to the LEOPARD. An interesting observation presents in the case of AM5$\rightarrow$AM4 where configuration (C) produces poor results. The performance betters in configuration (A) and (B) without the structural learning mechanism. Nevertheless, configuration (A)-(C) are not comparable to the LEOPARD where all modules are engaged. We note that the size of source stream is much less than the size of target stream in the AM5$\rightarrow$AM4 case. This leads to very few labelled samples provided in the warm-up phase. For AM1$\rightarrow$AM4, the absence of structural evolution, configuration (B), drops the LEOPARD's accuracy slightly. More severe degradation than configuration (B) occurs in configuration (A) and configuration (C) where the KL divergence loss and the cross domain adaptation loss are deactivated. The same pattern is demonstrated in the case of AM1$\rightarrow$AM3. This finding clearly confirms the advantage of the KL divergence loss and the cross domain adaptation loss for LEOPARD. In addition, the structural evolution boosts the performance of LEOPARD especially in the presence of drifts. The use of BERT as feature extractor as shown in Configuration (D) worsens predictive performances of LEOPARD significantly. This finding confirms the compatibility of the word2vec model over the BERT model as a feature extractor of LEOPARD most likely due to the absence of self-attention mechanism or recurrent connection in LEOPARD.  
    
	\begin{figure}[!t]
        \centering
        \subfloat[\centering before training]{{\includegraphics[width=3.5cm]{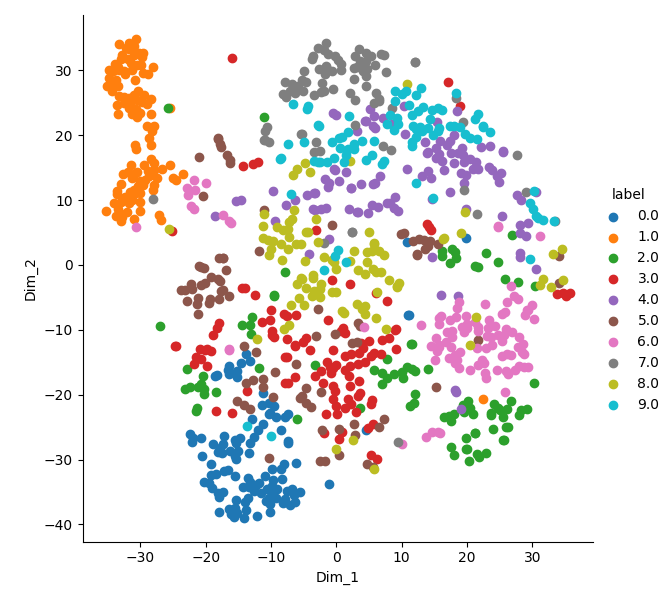} }}%
        \qquad
        \subfloat[\centering after  training]{{\includegraphics[width=3.5cm]{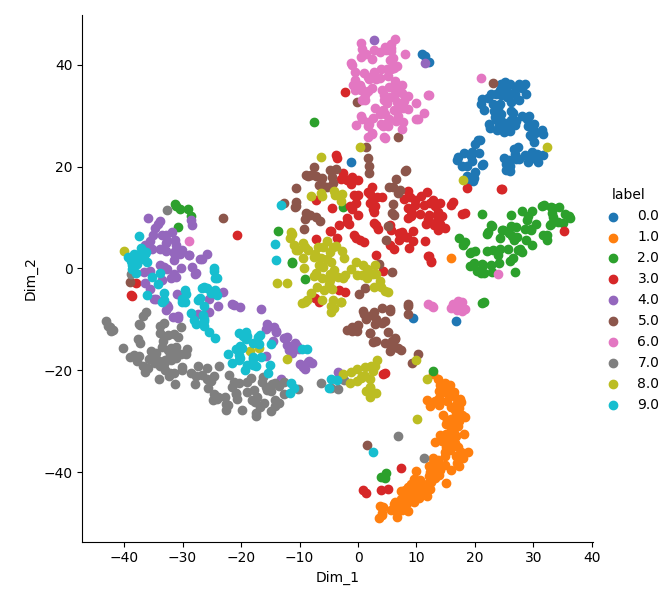} }}%
        \caption{USPS $\rightarrow$ MNIST tSNE plots on 1000 target data samples.}%
	    \label{fig:tSNE}
	\end{figure}
	
	\subsection{t-SNE Plots}
	Fig. \ref{fig:tSNE} illustrates the t-SNE plots of LEOPARD \cite{vandermaaten08a} for USPS$\rightarrow$MNIST case on the target stream before and after the training process. It is observed that there does not exist any cluster structures initially in this problem but such cluster structures are clearly present after the training process showing the effectiveness of \eqref{clusloss}. These facts confirm that LEOPARD does not call for the existence of any cluster structures beforehand and the clustering loss $L_{cluster}$ is capable of establishing the clustering-friendly latent space.
	
	\subsection{Future Directions}
	This paper has successfully developed an algorithmic solution of multistrean classification problems under extreme label shortages, LEOPARD. That is, given two different but related streaming processes, LEOPARD properly functions with few prerecorded samples of the source domain and the absence of any labels when the streaming processes run. This benefit goes one step ahead of existing multistream classifiers or unsupervised domain adaptation methods calling for fully labelled source streams. Nonetheless, the problem of multistream classification remain at the infant stages leaving several open issues for future works.
	
	The problem of open set domain adaptation presents a case where the source and target domains do not share the same target classes \cite{Busto2017OpenSD}. Such setting is also seen as a way to reduce the labelling cost and is beneficial in realm of multistream classifications. \cite{Busto2017OpenSD} proposes a feature transformation strategy associating target classes of target domain to those of source domain. \cite{Liu2019SeparateTA} puts forward the concept of openness and unknown classes in the open set domain adaptation problem. The theoretical bound of the open set domain adaptation is derived in \cite{Fang2021OpenSD}. The application of theoretical bound for deep learning is demontrated in \cite{Zhong2021BridgingTT}. These approaches are limited to the offline case calling for extensions for the multistream classification setting. Gradual Domain Adaptation \cite{Kumar2020UnderstandingSF} is highly relevant to the multistream classification context because the multistream classification problem still considers different but related streams and constant discrepancies. Also, the asynchronous drifts usually appear suddenly. \eqref{bound} assumes a small and fixed combined risk and is unrealistic because the combined risk may increase during the training process \cite{Zhong2021HowDT}. This issue is still ignored in the multistream classification problems. Few-shot Hypothesis adaptation \cite{Chi2021TOHANAO} is another interesting direction for the multistream classification topic and extends the few-shot domain adaptation problem without any source domain data.

	\section{Conclusion}
	 Learning Streaming Process from Partial Ground Truth (LEOPARD) is proposed in this paper to cope with the cross domain multistream classification problems under lack of labelled samples. The advantage of LEOPARD has been numerically validated using 24 study cases combined from nine datasets. It is demonstrated that LEOPARD outperforms its counterparts with noticeable margins in 15 of 24 cases. Ablation study further confirms the efficacy of LEOPARD's learning modules. One limitation of LEOPARD lies in the adversarial domain adaptation strategy only performing domain's alignment. This approach is poor when there exist big conditional distribution discrepancies. This issue is rather tricky here because LEOPARD does not benefit from any labels for its model updates. Our initial insight shows the feasibility of the pseudo-labelling strategy to attack this problem improving class inferences. Noisy labels remain an open issue and is explored in the future.
	 \section{Acknowledgement}
    We acknowledge the financial support of National Research Foundation, Singapore under IAFPP in the AME domain (contract no.: A19C1A0018) and the UniSA's start-up grant.
\bibliographystyle{IEEEtran}
\bibliography{bibliography}

\end{document}


\title{Supplemental Document of "Autonomous Cross Domain Adaptation under Extreme Label Scarcity"}

\author{Weiwei Weng, Mahardhika~Pratama,~\IEEEmembership{Senior Member,~IEEE}, Choiru Za'in, Marcus de Carvalho, \\Rakaraddi Appan, Andri Ashfahani, Edward Yapp Kien Yee
\thanks{W. Weng and M. Pratama share equal contributions. W. Weng, A. Ashfahani, M. de Carvalho, R. Appan are with the School of Computer Science and Engineering, Nanyang Technological University, Singapore. M. Pratama is with the academic unit of STEM, University of South Australia, Adelaide, Australia. C. Za'in is with the school of IT, Monash University. E. Y. Kien Yee is with Singapore Institute of Manufacturing Technology, A*Star, Singapore. Majority of the work was done when M. Pratama was with SCSE, NTU, Singapore. E-mail: weiwei.weng@ntu.edu.sg; dhika.pratama@unisa.edu.au;choiru.zain@monash.edu;marcus.decarvalho@
ntu.edu.sg;appan001@e.ntu.edu.sg;andriash001@e.ntu.edu.sg;edward\_yapp@
simtech.a-star.edu.sg.}}

\maketitle

\begin{abstract}
This document provides supplementary materials of our manuscript titled "Autonomous Cross Domain Adaptation under Extreme Label Scarcity". It contains six aspects: 1) the effect of label proportions $N_m$ is investigated; 2) visualization of LEOPARD's learning performance is offered; 3) hyper-parameters of all consolidated algorithms are reported in Table \ref{tab:param2} while Table \ref{tab:param1} presents the notations and symbols of the paper; 4) the use of MMD distance metric is investigated; 5) the MMD domain adaptation strategy is studied; 6) the pseudo labelling strategy is investigated.
\end{abstract}

\begin{IEEEkeywords}
Multistream Classification, Transfer Learning, Data Streams, Incremental Learning, Concept Drifts
\end{IEEEkeywords}

\begin{table*}[bp]
  \centering
  \caption{Analysis of Label's Proportions of Prerecorded Source Stream}
    \begin{tabular}{cccccc}
    \toprule
    Labeled proportion (\%) & 5     & 10    & 20    & 30    & 50 \\
    \midrule
    AM1 $\rightarrow$ AM3 & 71.07 $\pm$ 1.42 & 71.53 $\pm$ 1.08 & 70.17 $\pm$ 1.92 & 69.89 $\pm$ 2.63 & 72.53 $\pm$ 0.00 \\
    AM1 $\rightarrow$ AM4 & 57.49 $\pm$ 0.59 & 57.40 $\pm$ 0.54 & 57.87 $\pm$ 0.09 & 58.02 $\pm$ 0.17 & 57.82 $\pm$ 0.00 \\
    AM5 $\rightarrow$ AM1 & 74.42 $\pm$ 21.37 & 87.68 $\pm$ 0.38 & 84.21 $\pm$ 2.13 & 84.69 $\pm$ 1.92 & 84.53 $\pm$ 2.81 \\
    AM5 $\rightarrow$ AM4 & 45.74 $\pm$ 11.61 & 46.64 $\pm$ 8.03 & 46.17 $\pm$ 5.73 & 45.15 $\pm$ 6.99 & 45.83 $\pm$ 7.29 \\
    \bottomrule
    \end{tabular}%
  \label{tab:labeledrate}%
\end{table*}%

\section{Study of Label Proportions}
This section discusses the effect of label proportions $N_m$ to the performance of LEOPARD. Particularly, the label proportions for prerecorded source samples are varied to $5\%,10\%,20\%,30\%,50\%$ where our study is performed with four problems: AM1$\rightarrow$AM3, AM1$\rightarrow$AM4, AM5$\rightarrow$AM1 and AM5$\rightarrow$AM4. Table \ref{tab:labeledrate} reports our numerical results.

From Table \ref{tab:labeledrate}, it is observed that LEOPARD is not sensitive to the label's proportions. That is, the gap in performance is not significant given varied label proportions. For AM1$\rightarrow$AM3 case, LEOPARD's accuracy slightly betters by less than $1\%$ with a $50\%$ label quantity. On the other hand, its performance does not degrade drastically with a $5\%$ label proportion. AM1$\rightarrow$AM4 shows stable performances of LEOPARD in around $57\%$ regardless of label's proportions. The same pattern is also observed in the AM5$\rightarrow$AM4 case where the accuracy is around 46\%. A significant deterioration presents in the case of AM5$\rightarrow$AM1 where around 13\% reduction in performance exists with a 5\% label quantity. This case occurs because a lack of labelled samples causes inaccurate cluster's allegiance measuring the cluster-to-class relationship. On the other hand, the increase of label performances does not lead to performance's improvements in most other cases because these labelled samples may not contain important label information. Note that labelled samples are randomly selected from a uniform distribution here. 

\begin{table*}
    \centering
    \caption{Symbols \& Notations}
    \begin{tabular}{l|l}
    \toprule
        Symbol & Description \\
        \midrule
        $x_{S,T}$           & Raw input \\
        $Z_{S,T}$           & Latent input features \\
        $D_S, D_T$          & Source domain and target domain\\
        $B_{k_s}^S,B_{k_t}^T$    & $k_s-th (k_t-th)$ batch of source data stream (target data stream) \\
        $W_{enc}^l$         & Connective weights of the encoder \\
        $W_{dec}^l$         & Connective weights of the decoder \\
        $b^l$               & Biases of the encoder \\
        $c^l$               & Biases of the decoder \\
        $h^l$               & Latent representation of the $l-th$ hidden layer \\
        $C_j^l$             & Centroid of $j-th$ cluster of $l-th$ layer \\
        $Clus^l$            & Cluster number in the $l-th$ hidden layer \\
        $Ale_{j,o}^l$       & Cluster allegiance \\
        $Score^l$           & Local score of the $l-th$ layer \\
        $\hat{Y}$           & Predicted class label \\
        $d_{H}(\cdot,\cdot)$& $H$ divergence \\
        $\zeta_{W_{DC}}(\cdot)$ & Domain classifier \\
        $\xi_{W_{C}}(\cdot)$ & Classifier \\
        $F_{W_f}(\cdot)$        & Feature extractor \\
        $L_{cluster}$       & Clustering loss \\
        $L_{cd}$            & Cross-domain adaptation loss \\
        $L_{\zeta}(\cdot)$          & Cross entropy loss function \\
        $L_{\xi}(\cdot)$    & Reconstruction (mean square error) loss function \\
        $KL(\phi^l|\Phi^l)$ & KL divergence loss \\
        $\phi^l_j$          & Similarity degree of the $j-th$ cluster of $l-th$ layer \\
        $\Phi^l$            & Auxiliary target distribution of the $l-th$ latent space \\
        $d_n$               & Origin of data sample \\
        $\zeta_j$           & Frequency of a cluster \\
        $D(\cdot,\cdot)$    & $L_2$ distance between two variables \\
        $\mu_{D,i}^l$       & Mean of $D(h^l,C_i^l)$ \\
        $\sigma_{D,i}^l$    & Standard deviation of $D(h^l,C_i^l)$ \\
        $k_1$               & Dynamic confident degree \\
        $\epsilon_{D,W}$    & Error bounds \\
        $E$                 & Epoch number \\
    \bottomrule
    \end{tabular}
    \label{tab:param1}
\end{table*}

\begin{table*}
\footnotesize
    \centering
    \caption{Hyper-parameters of each algorithm}
    \begin{tabular}{l|l|l}
    \toprule
        Symbol & Value & Description \\
        \midrule
        $\lambda$ & 1 & used in the definition of data sample similarity degree \\
        $\alpha_1$ & 0.1 & cross-domain adaptation loss trade-off constant \\
        $\alpha_2$ & 1.0 & KL divergence regularization constant \\
        \multirow{4}[0]{*}{$E_{clus}$} & {$Amazon$@X: 1} & \multirow{4}[0]{*}{epochs for hidden layer representation updates} \\
          & $MNIST\rightarrow USPS$: 5 &  \\
          & $USPS\rightarrow MNIST$: 10 &  \\
          & ${Office31}$: 10 & \\
        $E_{dc}$ & 1 & epochs for updating cross domain adaptation \\
        $\alpha_x$ & 0.001 & the significance level applying Hoeffding’s bound \\
        $\alpha_W$ & 0.001 & drift warning condition: lower bound \\
        $\alpha_D$ & 0.005 & drift warning condition: upper bound \\
        
    \bottomrule
    \end{tabular}
    \label{tab:param2}
\end{table*}

\begin{figure}
    \centering
    \includegraphics[width=0.45\textwidth]{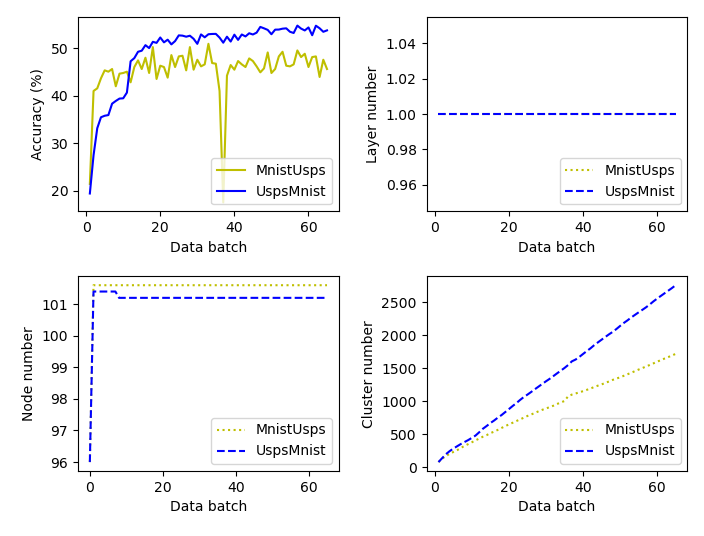}
    \caption{LEOPARD performance of MNIST$\rightarrow$USPS (yellow) and USPS$\rightarrow$MNIST (blue) problem (Upper left: target accuracy trajectory along with the number of data batch; upper right: layer evolution of LEOPARD classifier; lower left: node evolution; lower right: the trace of cluster number).}
    \label{fig:mnist-usps}
\end{figure}

\begin{figure}
    \centering
    \includegraphics[width=0.45\textwidth]{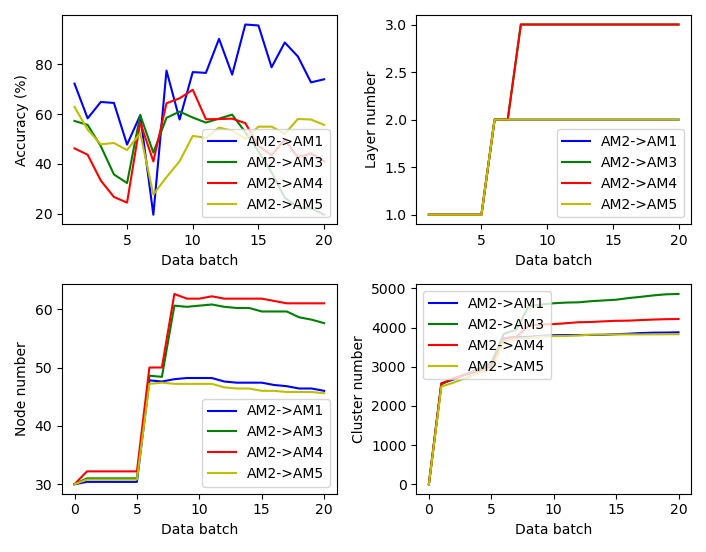}
    \caption{LEOPARD performance of Amazon@Books (Source dataset) experiments (Upper left: target accuracy trajectory along with the number of data batch; upper right: layer evolution of LEOPARD classifier; lower left: node evolution; lower right: the trace of cluster number).}
    \label{fig:amazon@book}
\end{figure}
\begin{table*}[htbp]
\scriptsize
  \centering
  \caption{Additional Ablation Study of LEOPARD}
  \begin{threeparttable}
    \begin{tabular}{ccccccccc}
    \toprule
          & A     & B     & C  & D & E & F & G & LEOPARD \\
    \midrule
    AM1 $\rightarrow$ AM3 & 71.48 $\pm$ 0.78 & 67.89 $\pm$ 8.06 & 71.01 $\pm$ 1.04 & 45.83 $\pm$ 8.28 & 45.20 $\pm$ 7.60 & 70.23 $\pm$ 1.68 & \textbf{72.53 $\pm$ 0.00} & {71.53 $\pm$ 1.08} \\
    AM1 $\rightarrow$ AM4 & 57.78 $\pm$ 0.01 & 57.78 $\pm$ 0.03 & {57.79 $\pm$ 0.02} & 38.40 $\pm$ 3.13 & 51.14 $\pm$ 2.81 & {57.80 $\pm$ 0.02} & \textbf{57.82 $\pm$ 0.00} & {57.78 $\pm$ 0.05} \\
    AM5 $\rightarrow$ AM1 & {87.96 $\pm$ 0.26} & 87.40 $\pm$ 1.01 & 85.59 $\pm$ 3.89 & 43.52 $\pm$ 16.20 & 85.87 $\pm$ 2.40 & 38.92 $\pm$ 12.24 & \textbf{88.29 $\pm$ 0.00} & {87.68 $\pm$ 0.38} \\
    AM5 $\rightarrow$ AM4 & 49.06 $\pm$ 7.79 & 49.53 $\pm$ 4.32 & 41.19 $\pm$ 10.63 & 26.22 $\pm$ 8.55 & 31.21 $\pm$ 4.38 & 42.72 $\pm$ 10.32 & 50.52 $\pm$ 4.67 & \textbf{50.76 $\pm$ 2.70} \\
        \bottomrule
    \end{tabular}%
    \begin{tablenotes}
          \item[A] MMD with Gaussian kernel in Clustering-Friendly Latent Space
          \item[B] MMD with rbf kernel in Clustering-Friendly Latent Space
          \item[C] MMD with multiscale kernel in Clustering-Friendly Latent Space
          \item[D] MMD-D: MMD with a deep kernel whose parameters are optimized
          \item[E] MMD-O: MMD with a Gaussian kernel whose lengthscale is optimized
          \item[F] MMD-O + domain discriminator
          \item[G] $10\%$ pseudo labelled target data
    \end{tablenotes}

    \end{threeparttable}
  \label{ablation2}%
\end{table*}%

\section{Visualization of LEOPARD Learning Performance}
This section offers the visual illustration of LEOPARD learning performances. Fig. \ref{fig:mnist-usps} portrays the evolution of LEOPARD's hidden nodes, layers and clusters as well as the trace of classification rates in the target stream. These figures are produced using the MNIST$\rightarrow$USPS and USPS$\rightarrow$MNIST problem. Fig. \ref{fig:amazon@book} depicts the evolution of LEOPARD's hidden nodes, hidden layers, hidden clusters and classification rates for the target stream in the amazon@books problems. These plots delineate an average of 5 independent numerical results. 

From Fig. \ref{fig:mnist-usps}, the self-evolving mechanism of LEOPARD is demonstrated. For USPS$\rightarrow$MNIST problem, several nodes are added in the beginning of the training process followed by the pruning process of inactive nodes. Hidden nodes are stable afterward. Note that LEOPARD starts from 96 nodes. The asynchronous drift occurs at $k=35$ for the source stream and at $k=36$ for the target stream which do not affect LEOPARD's learning performances significantly. This condition is expected because LEOPARD grows the hidden clusters and ends up with around 3000 clusters. Note that the deep clustering approach only performs well with high numbers of clusters. For MNIST$\rightarrow$USPS problem, it is perceived that the asynchronous drifts significantly undermines LEOPARD's classification performance. However, LEOPARD quickly recovers from this condition meaning that the asynchronous drifts are successfully compensated. The hidden node growing process takes place in the beginning. This situation happens because LEOPARD is still in the high bias situation, under-fitting.  

From Fig. \ref{fig:amazon@book}, the performances of LEOPARD compromises due to the asynchronous drift at $k=5$ and $k=6$ but successfully recovers showing improved overall trends. LEOPARD reacts to the concept drifts with introductions of new nodes and new layers. The hidden clusters increases exponentially in the beginning and is stable afterward at around $3000-5000$ clusters. LEOPARD features an open structure which dynamically evolves its hidden nodes, hidden clusters and hidden layers. This mechanism possesses dual advantages where it automatically generates desirable network structures for the given problems and compensates the concept drifts.

\section{The MMD distance metric}
This section explores the application of the MMD distance metric instead of the KL divergence measure in the clustering loss $L_{cluster}$ of (14). As with the ablation study section and the analysis of label proportions section, four study cases, AM1$\rightarrow$ AM3, AM1$\rightarrow$ AM4, AM5$\rightarrow$ AM1, AM5$\rightarrow$AM4, are put forward where numerical results with three MMD kernels: Gausssian, RBF and multiscale, are presented in Table \ref{ablation2}. There do not exist any significant differences among the four LEOPARD configurations because both the KL divergence measure and the MMD distance measure function similarly. Note that the goal here is to inform the difference between the current distribution and the target distribution to be minimized using the SGD optimization approach. We also expect the same finding to be returned with other distance measures.

\section{MMD Domain Adaptation Strategy}
This section simulates the MMD domain adaptation approach instead of the adversarial domain adaptation method as used in the LEOPARD learning strategy. Three configurations are attempted here, MMD-D, MMD-O and MMD-O+domain discriminator, i.e., the MMD approach is applied alongside with the adversarial domain adaptation strategy. These three configurations are presented in Table \ref{ablation2} as D, E, F. Table \ref{ablation2} confirms that LEOPARD is the best-performing algorithm compared to those other three algorithms. Configuration D where the MMD approach with a deep kernel is used leads to major performance deterioration, over $20\%$ drops compared to the original LEOPARD. Configuration E where the MMD approach with the Gaussian kernel is utilized is relatively better than the configuration E but still worse than LEOPARD. On the other hand, configuration F combining the MMD-O and the adversarial domain adaptation significantly improves from configuration E and F but remains lower than LEOPARD. This finding confirms the advantage of the adversarial domain adaptation compared to the MMD approach.

\section{Pseudo-Labelling Strategy}
This section explores the pseudo-labelling strategy to improve the performance of LEOPARD. Note that LEOPARD utilizes the adversarial domain adaptation approach where only the marginal distribution minimization is implemented. This strategy is chosen because LEOPARD does not utilize any labels for its model updates. We apply the pseudo-labelling strategy here to improve the performance of LEOPARD further. Configuration F of Table \ref{ablation2} reports our numerical results. It is seen from Table \ref{ablation2} that some performance gain is achieved for three cases $AM1\rightarrow AM3, AM1\rightarrow AM4, AM5\rightarrow AM1$ but performance drop is observed for $AM5\rightarrow AM4$. This performance drop is expected because of the presence of noisy labels. The performance improvement is also not significant using pseudo labels because no labels are utilized for model updates, i.e., labels are only fed for class inferences.
